\documentclass[10pt,twocolumn,letterpaper]{article}

\usepackage{cvpr}
\usepackage{times}
\usepackage{epsfig}
\usepackage{graphicx}
\usepackage{amsmath}
\usepackage{amssymb}
\usepackage{diagbox}
\usepackage{authblk}

\usepackage[pagebackref=true,breaklinks=true,letterpaper=true,colorlinks,bookmarks=false]{hyperref}
\usepackage{color}

\cvprfinalcopy 

\def\cvprPaperID{323} 

\begin{document}

\setlength{\belowdisplayskip}{7pt} \setlength{\belowdisplayshortskip}{7pt}
\setlength{\abovedisplayskip}{0pt} \setlength{\abovedisplayshortskip}{0pt}

\title{Semantic-aware Grad-GAN for Virtual-to-Real Urban Scene Adaption}
\author[1]{Peilun Li}
\author[1]{Xiaodan Liang}
\author[1]{Daoyuan Jia}
\author[2]{Eric P. Xing}
\affil[ ]{\textit {\{peilunl@andrew,xiaodan1@cs,daoyuanj@andrew,epxing@cs\}.cmu.edu}}
\affil[1]{School of Computer Science, Carnegie Mellon University }
\affil[2]{Petuum, Inc.}

\maketitle

\begin{abstract}
Recent advances in vision tasks (\eg, segmentation) highly depend on the availability of large-scale real-world image annotations obtained by cumbersome human labors. Moreover, the perception performance often drops significantly for new scenarios, due to the poor generalization capability of models trained on limited and biased annotations. In this work, we resort to transfer knowledge from automatically rendered scene annotations in virtual-world to facilitate real-world visual tasks. Although virtual-world annotations can be ideally diverse and unlimited, the discrepant data distributions between virtual and real-world make it challenging for knowledge transferring. We thus propose a novel Semantic-aware Grad-GAN (SG-GAN) to perform virtual-to-real domain adaption with the ability of retaining vital semantic information. Beyond the simple holistic color/texture transformation achieved by prior works, SG-GAN successfully personalizes the appearance adaption for each semantic region in order to preserve their key characteristic for better recognition. It presents two main contributions to traditional GANs: 1) a soft gradient-sensitive objective for keeping semantic boundaries; 2) a semantic-aware discriminator for validating the fidelity of personalized adaptions with respect to each semantic region. Qualitative and quantitative experiments demonstrate the superiority of our SG-GAN in scene adaption over state-of-the-art GANs. Further evaluations on semantic segmentation on Cityscapes show using adapted virtual images by SG-GAN dramatically improves segmentation performance than original virtual data. We release our code at \url{https://github.com/Peilun-Li/SG-GAN}.
\end{abstract}

\section{Introduction} \label{Intro}

\begin{figure*}
\centering
\begin{tabular}{c} 
\includegraphics[width=0.89\linewidth]{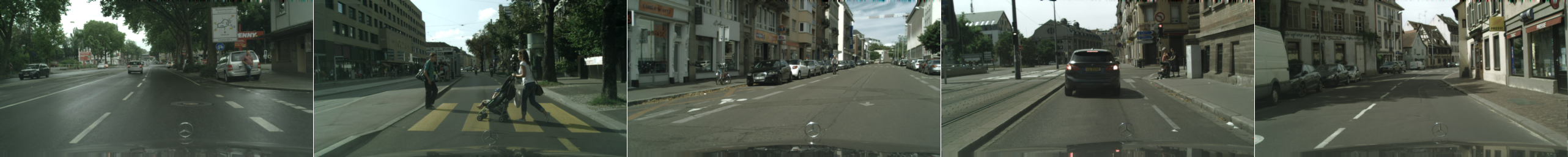}\\
\small{(a) Real-world (Cityscapes)}\\
\includegraphics[width=0.89\linewidth]{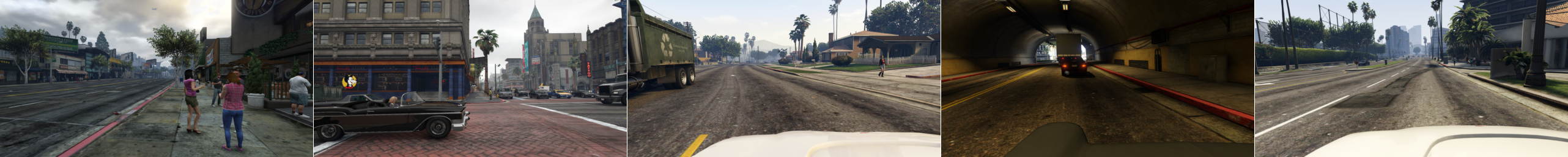}\\
\small{(b) Virtual-world (GTA-V)}\\
\end{tabular}
   \caption{Visual comparison between real-world images and virtual-world images. (a) Real-world images sampled from Cityscapes dataset \cite{Cityscapes}. (b) Virtual-world images sampled from GTA-V dataset \cite{Playing}. 
   }\vspace{-6mm}
\label{fig:virtual-vs-real}
\end{figure*}

Recently, very promising visual perception performances on a variety of tasks (\eg classification and detection) have been achieved by deep learning models \cite{resnet,hu2017squeeze,huang2017densely,ren2015faster}, driven by the large-scale annotated datasets. However, more fine-grained tasks (\eg semantic segmentation) still have much space to be resolved due to the insufficient pixel-wise annotations in diverse scenes. High quality annotations are often prohibitively difficult to obtain with the need of tons of human efforts, \eg, Cityscapes dataset \cite{Cityscapes} reports it will take more than 90 minutes for manually labeling a single image. Moreover, the learned models on limited and biased dataset often tend to not generalize well on other datasets in different domains, as demonstrated in prior domain adaption works \cite{hoffman2016fcns}. 

An alternative solution to alleviate this data issue is to seek an automatic data generation approach. Rather than relying on expensive labors on annotating real-world data, recent progresses in Computer Graphics \cite{Driving,PlayingForBenchmarks,Playing} make it possible to automatically or semi-automatically capture both images and their corresponding semantic labeling from video games, \eg, Grand Theft Auto V (GTA V), which is a realistic open-world game based on Los Angeles. In virtual-world, we can easily collect diverse labeled data that is several orders of magnitude larger than real-world human annotations in an unlimited way.

However, utilizing virtual-world knowledge to facilitate real-world perception tasks is not a trivial technique due to the common severe domain shift problem \cite{quionero2009dataset}. Images collected from virtual-world often yield inconsistent distribution with that of real-world ones, because of the limitation in rendering and object simulation techniques, as shown in Figure \ref{fig:virtual-vs-real}. It is thus desirable to bridge the gap between virtual-world and real-world data for exploiting the shared semantic knowledge for perception. Previous domain adaption approaches can be summarized as two lines: minimizing the difference between the source and target feature distributions~\cite{Gebru_2017_ICCV, hoffman2014lsda, hoffman2015detector, hoffman2016large,hoffman2016fcns,Adda_CVPR2017}; or explicitly ensuring that two data distributions close to each other by adversarial learning~\cite{discogan,unit,liu2016coupled,SimGAN,dualgan,CycleGAN} or feature combining~\cite{NeuralStyle, gatys2016preserving, perceptual, MMD, ulyanov2016texture}. On the one hand, those feature-based adaption methods require the supervision for each specific task in both source and target domains, which cannot be widely applicable. On the other hand, despite the promising adaption performance achieved by Generative Adversarial Networks (GANs) \cite{GAN}, where a discriminator is trained to distinguish fake images from real images and a generator is optimized for generating realistic images to deceive discriminator, existing models can only transfer holistic color and texture of the source images to target images while disregarding the key characteristics of each semantic region (\eg road vs. car), yielding very blurry and distorted results. The loss of fine-grained details in generated images would severely hinder their capabilities of facilitating downstream vision perception tasks. 

In this work, we propose a novel Semantic-aware Grad-GAN (SG-GAN) that aims at transferring personalized styles (\eg color, texture) for distinct semantic regions in virtual-world images to approximate the real-world distributions. Our SG-GAN, as one kind of image-based adaption approaches, is able to not only preserve key semantic and structure information in source domain but also enforce each semantic region close to their corresponding real-world distributions. 

Except the traditional adversarial objective used in prior GANs, we propose two main contributions to achieve the above mentioned goals. First, a new gradient-sensitive objective is introduced into optimizing the generator, which emphasizes the semantic boundary consistencies between virtual images and adapted images. It is able to regularize the generator render distinct color/texture for each semantic region in order to keep semantic boundaries, which can alleviate the common blurry issues. Second, previous works often learn a whole image discriminator for validating the fidelity of all regions, which makes the color/texture of all pixels in original images easily collapse into a monotonous pattern. We here argue that the appearance distributions for each semantic region should be regarded differently and purposely. For example, \emph{road} region in real-world often appears with coarse texture of asphalt concrete while \emph{vehicle} region is usually smooth and reflective. In contrast to standard discriminator that eventually examines on a global feature map, we employ a new semantic-aware discriminator for evaluating the image adaption quality in a semantic-wise manner. The semantic-aware discriminator learns distinct discriminate parameters for examining regions with respect to each semantic label. This distinguishes SG-GAN with existing GANs as a controllable architecture that personalizes texture rendering for different semantic regions and results in adapted images with finer details.

Extensive qualitative and quantitative experiments on adapting GTA-V virtual images demonstrate that our SG-GAN can successfully generate realistic images without changing semantic information. To further demonstrate the quality of adapted images, we use the adapted images to train semantic segmentation models and evaluate them on public Cityscapes dataset \cite{Cityscapes}. The substantial performance improvement over using original virtual data on semantic segmentation speaks well the superiority of our SG-GAN for semantic-aware virtual-to-real scene adaption. 



\section{Related work} \label{Related}
\textbf{Real-world vs. virtual-world data acquiring:} Fine-grained semantic segmentation on urban scenes takes huge amount of human effort, which results in much less data than that of image classification datasets, as referred to as ``curse of dataset annotation" in \cite{Curse}. For example, CamVid dataset \cite{CamVid} provides 700 road scene images with an annotation speed of 60 minutes/image. Cityscapes dataset \cite{Cityscapes} releases 5000 road scene annotations and reports annotation speed as more than 90 minutes/image. On the contrary, collecting urban scene data from video games such as GTA V has attracted lots of interests \cite{Driving,PlayingForBenchmarks,Playing} for automatically obtaining a large amount of data. Specifically, Richter \etal \cite{Playing} inject the connection between GTA V and GPU to collect rendered data and develop an interactive interface to extract 24966 images with annotations within 49 hours. Richter \etal \cite{PlayingForBenchmarks} further develop real-time rendering pipelines enabling video-rate data and groundtruth collection, and release a dataset of 254064 fully annotated video frames. However, despite its diversity, virtual-world scene data often looks very unrealistic (\eg flawed lighting and shadowing) due to the imperfect texture rendering. Directly utilizing such unrealistic data would damage real-world visual tasks due to their discrepant data distributions.


\textbf{Domain adaption:} Domain adaption can be approached by either adapting scene images or adapting hidden feature representations guided by the targets. Image-based adaption can be also referred to as image-to-image translation, i.e., translating images from source domain to target domain, which can be summarized into two following directions. 

First, adapted images can be generated through feature matching \cite{NeuralStyle, gatys2016preserving, perceptual, MMD, ulyanov2016texture}. Gatys \etal \cite{NeuralStyle} propose a method to combine content of one image and style of another image through matching Gram matrix on deep feature maps, at the expense of some loss of content information. Second, a generative model can be trained through adversarial learning for image translation. Isola \etal \cite{pix2pix} use conditional GANs to learn mapping function from source domain to target domain, with a requirement of paired training data, which is unpractical for some tasks. To remove the requirement of paired training data, extra regularization could be applied, including self-regularization term \cite{SimGAN} , cycle structure \cite{discogan, dualgan, CycleGAN} or weight sharing \cite{unit,liu2016coupled}. There are also approaches making use of both feature matching and adversarial learning \cite{cha2017adversarial,xiong2017learning}. However, in urban scene adaption, despite having the ability to generate relatively realistic images, existing approaches often modify semantic information, \eg, the sky will be adapted to tree structure, or a road lamp may be rendered from nothing. 

In contrast to image-based adaption that translates images to target domain, hidden feature representation based adaption aims at adapting learned models to target domain  \cite{Gebru_2017_ICCV,hoffman2014lsda,
hoffman2016large,hoffman2016fcns, liang2017recurrent,liang2017dual,liang2017generative,
Adda_CVPR2017}. By sharing weight \cite{Gebru_2017_ICCV} or incorporating adversarial discriminative setting \cite{Adda_CVPR2017}, those feature-based adaption methods help mitigate performance degradation caused by domain shifting. However, feature-based adaption methods require different objective or architecture for different vision tasks, thus not as widely-applicable as image-based adaption.

\textbf{Image synthesis:} Apart from domain adaption, there exist some other approaches that generate realistic image from text \cite{reed2016generative, han2017stackgan}, segmentation groundtruth \cite{CRN} and low resolution image \cite{bansal2017pixelnn}. For instance, cascaded refinement network \cite{CRN} is proposed to synthesize realistic images from semantic segmentation input, and semantic information can be preserved as our approach. However, since semantic segmentation is by nature much harder to be retrieved than raw image, image translation approaches has more potential in large scale data generation. 

\begin{figure}[t]
\begin{center}
\includegraphics[width=0.8\linewidth]{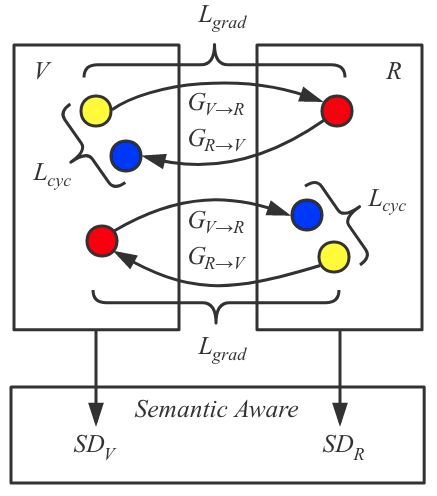}\vspace{-4mm}
\end{center}
   \caption{Illustration of the proposed Semantic-aware Grad-GAN (SG-GAN). We represent the unpaired virtual-world images and real-world images as yellow dots in $V$ and $R$ boxes, respectively. Two symmetric generators $G_{V\rightarrow R}$, $G_{R\rightarrow V}$ are learned to perform scene adaption from each other. In addition to the cycle consistency loss \cite{CycleGAN}, a new soft gradient-sensitive objective $L_{grad}$ is imposed on generators for ensuring semantic boundaries of the original image and its adapted image pairs be consistent. Two semantic-aware discriminators $\text{SD}_V$ and $\text{SD}_R$ are jointly optimized for examining the fidelity of adapted real-world images and virtual-world images, respectively.}\vspace{-6mm}
\label{fig:arch}
\end{figure}

\section{Semantic-aware Grad-GAN} \label{Main}

The goal of the proposed SG-GAN is to perform virtual-to-real domain adaption while preserving their key semantic characteristics for distinct contents. Capitalized on the Generative Adversarial Networks (GANs), SG-GAN presents two improvements over the traditional GAN model, i.e., a new soft gradient-sensitive objective over generators and a novel semantic-aware discriminator.

\subsection{Semantic-aware cycle objective}

Our SG-GAN is based on the cycle-structured GAN objective since it has shown the advantages of training stability and generation quality~\cite{discogan, dualgan, CycleGAN}. Specifically, let us denote the unpaired images from the virtual-world domain $V$ and real-world domain $R$ as $\{v\}_{i=1}^N \in V$ and $\{r\}_{j=1}^M \in R$, respectively. Our SG-GAN learns two symmetric mappings $G_{V\rightarrow R}$, $G_{R\rightarrow V}$ along with two corresponding semantic-aware discriminators $\mathit{SD_R}$, $\mathit{SD_V}$ in an adversarial way. $G_{V\rightarrow R}$ and $G_{R\rightarrow V}$ map images between virtual-world and real-world domains. $\mathit{SD_R}$'s target is to distinguish between real-world images $\{r\}$ and fake real-world images $\{G_{V\rightarrow R}(v)\}$, and vice versa for $\mathit{SD_V}$. The details of semantic-aware discriminators will be introduced later in Section \ref{sec:sd}. Figure \ref{fig:arch} illustrates the relationship of $V$, $R$, $G_{V\rightarrow R}$, $G_{R\rightarrow V}$, $\mathit{SD_V}$ and $\mathit{SD_R}$.

\subsubsection{Adversarial loss}

\indent

Our objective function is constructed based on standard adversarial loss \cite{GAN}. Two sets of adversarial losses are applied to $(G_{V\rightarrow R}, \mathit{SD_R})$ and $(G_{R\rightarrow V}, \mathit{SD_V})$  pairs. Specifically, the adversarial loss $L_{adv}$ for optimizing $(G_{V\rightarrow R}, \mathit{SD_R})$ is defined as:

\begin{equation}\label{eq:adv}
\begin{aligned}
&L_{adv}(G_{V\rightarrow R},\mathit{SD_R},V,R)\\
=&\mathbb{E}_{r\sim p_{data}(r)}[\log \mathit{SD_R}(r)] \\
+& \mathbb{E}_{v\sim p_{data}(v)}[\log (1-\mathit{SD_R}(G_{V\rightarrow R}(v))]
\end{aligned}
\end{equation}

Note that this is a mini-max problem as $\mathit{SD_R}$ aims to maximize $L_{adv}$ and $G$ aims to minimize $L_{adv}$. The objective of $G_{V\rightarrow R}^*$ can be formulated as:

\begin{equation}
\begin{aligned}
G_{V\rightarrow R}^* = \arg \min_{G_{V\rightarrow R}} \max_{\mathit{SD_R}} L_{adv}(G_{V\rightarrow R},\mathit{SD_R},V,R)
\end{aligned}
\end{equation}

The formula is similar for the generator $G_{R\rightarrow V}$ and semantic-aware discriminator $\mathit{SD_V}$, of which the adversarial loss can be noted as $L_{adv}(G_{R\rightarrow V},\mathit{SD_V},R,V)$.

\subsubsection{Cycle consistency loss}

\indent

Another part of our objective function is cycle consistency loss \cite{CycleGAN}, which is shown helpful to reduce the space of possible mappings, i.e., $G_{V\rightarrow R}$ and $G_{R\rightarrow V}$. The cycle consistency loss confines that after going through $G_{V\rightarrow R}$ and  $G_{R\rightarrow V}$, an image should be mapped as close as to itself, i.e., $G_{R\rightarrow V}(G_{V\rightarrow R}(v))\approx v$, $G_{V\rightarrow R}(G_{R\rightarrow V}(r))\approx r$. In this work, we define cycle consistency loss as:

\begin{equation}
\begin{aligned}
&L_{cyc}(G_{V\rightarrow R},G_{R\rightarrow V},V,R)\\
=&\mathbb{E}_{r\sim p_{data}(r)}[||G_{V\rightarrow R}(G_{R\rightarrow V}(r))-r||_1]\\
+& \mathbb{E}_{v\sim p_{data}(v)}[||G_{R\rightarrow V}(G_{V\rightarrow R}(v))-v||_1]
\end{aligned}
\end{equation}

Cycle consistency loss can be seen as introducing a regularization on positions of image elements. Mapping functions are trained in a way that moving positions of image components is not encouraged. However, as position is only a fraction of semantic information, cycle consistency loss itself can't guarantee to preserve well semantic information. For complex adaption such as urban scene adaption, a model purely with cycle consistency loss often fails by wrongly mapping a region with one semantic label to another label, \eg, the sky region may be wrongly adapted into a tree region, as shown in Figure \ref{fig:vis}. This limitation of cycle structure is also discussed in \cite{CycleGAN}.

\subsubsection{Soft gradient-sensitive objective}

\indent

In order to keep semantic information from being changed through the mapping functions, we introduce a novel soft gradient-sensitive loss, which uses image's semantic information in a gradient level. We first introduce gradient-sensitive loss, and then show ways to make the gradient-sensitive loss into a soft version. 

The motivation of gradient-sensitive loss is that no matter how texture of each semantic class changes, there should be some distinguishable visual differences at the boundaries of semantic classes. Visual differences for adjacent pixels can be captured through convolving gradient filters upon the image. A typical choice of gradient filter is Sobel filter \cite{Sobel} as $\mathbf{C}=\{C_x,C_y\}$ defined in Equation \ref{eq:sobel}. 


\begin{equation} \label{eq:sobel}
\begin{aligned}
C_x = \begin{pmatrix}
-1 & 0 & 1\\
-2 & 0 & 2\\
-1 & 0 & 1\\
\end{pmatrix}&,
C_y = \begin{pmatrix}
1 & 2 & 1\\
0 & 0 & 0\\
-1 & -2 & -1\\
\end{pmatrix}\\
\end{aligned}
\end{equation}

Since our focus is visual differences on semantic boundaries, a 0-1 mask is necessary that only has non-zero values on semantic boundaries. Such mask can be retrieved by convolving a gradient filter upon semantic labeling since it only has different adjacent values on semantic boundaries. Semantic labeling can be obtained by human annotation, segmentation models \cite{ademxapp}, or Computer Graphics tools \cite{Driving,PlayingForBenchmarks,Playing}. By multiplying the convolved semantic labeling and the convolved image element-wise, attention will only be paid to visual differences on semantic boundaries.

More specifically, for an input image $v$ and its corresponding semantic labeling $s_v$, since we desire $v$ and $G_{V\rightarrow R}(v)$ share the same semantic information, the gradient-sensitive loss for image $v$ can be defined as Equation \ref{eq:gl}, in which $\mathbf{C}_i$ and $\mathbf{C}_s$ are gradient filters for image and semantic labeling, $*$ stands for convolution, $\odot$ stands for element-wise multiplication, $|\cdot|$ represents absolute value, $||\cdot||_1$ means L1-norm, and $sgn$ is the sign function.

\begin{equation} \label{eq:gl}
\begin{aligned}
&l_{grad}(v,s_v,G_{V\rightarrow R}) \\
=& ||(|(|\mathbf{C}_i*v| - |\mathbf{C}_i*G_{V\rightarrow R}(v)|)|)\\ 
\odot& sgn(\mathbf{C}_s*s_v)||_1
\end{aligned}
\end{equation}

\begin{figure*}
\begin{center}
\includegraphics[width=1\linewidth]{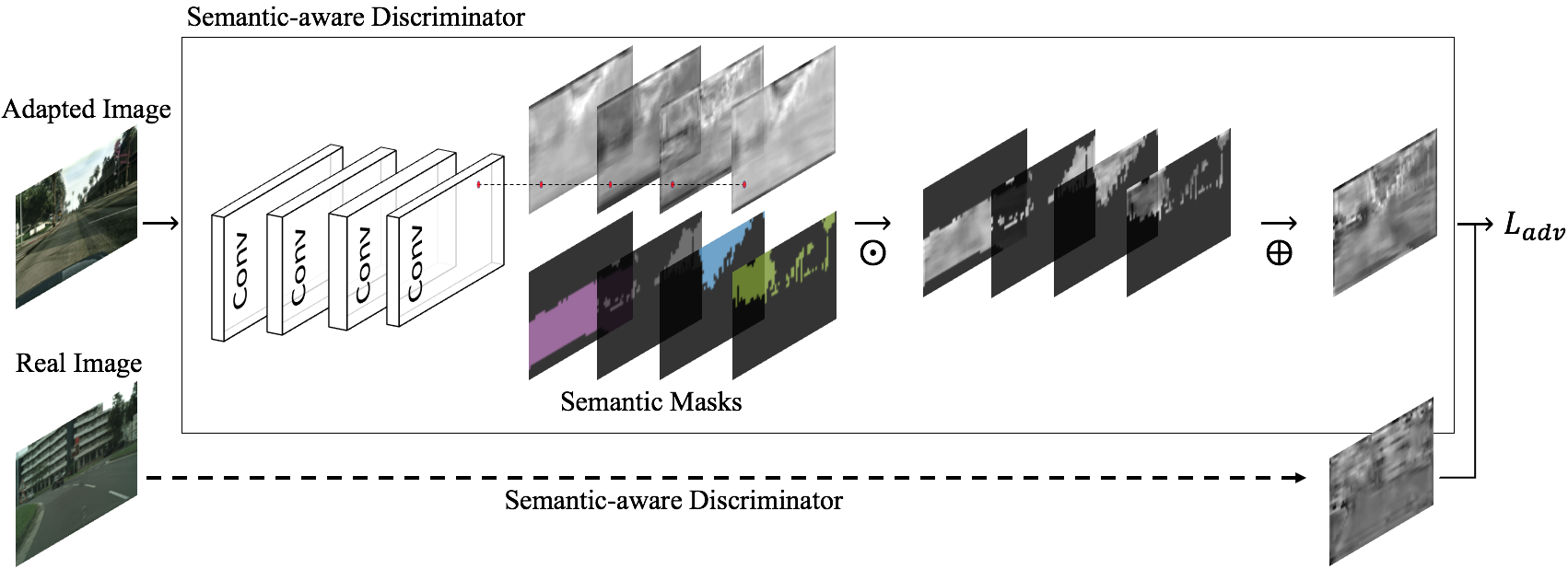}\vspace{-4mm}
\end{center}
   \caption{Illustration of semantic-aware discriminator which takes either real image or adapted image as inputs and is then optimized with an adversarial objective. Each input is first passed through several convolution layers and the resulting feature maps are multiplied with semantic masks element-wisely, and then summed into a single channel output. The coupled outputs are used for optimizing adversarial loss as Equation \ref{eq:adv}. $\odot$ represents the element-wise multiplication operation while $\oplus$ as the summation operation over the channel dimension. The sampled feature maps are rescaled to [0,255] for better visualization.}\vspace{-6mm}
\label{fig:dis}
\end{figure*}

In practice, we may hold belief that $v$ and $G_{V\rightarrow R}(v)$ share similar texture within semantic classes. Since texture information can also be extracted from image gradient, a soft gradient-sensitive loss for image $v$ can be defined as Equation \ref{eq:wgl} to represent such belief, in which $\beta$ controls how much belief we have on texture similarities. 

\begin{equation} \label{eq:wgl}
\begin{aligned}
&l_{s-grad}(v,s_v,G_{V\rightarrow R},\alpha,\beta) \\
=& ||(|(|\mathbf{C}_i*v| - |\mathbf{C}_i*G_{V\rightarrow R}(v)|)|)\\ 
\odot& (\alpha \times |sgn(\mathbf{C}_s*s_v)| + \beta)||_1\\
&s.t.\ \  \alpha+\beta=1\ \  \alpha,\beta\geq 0\\
\end{aligned}
\end{equation}

Given the soft gradient-sensitive loss for a single image, the final objective for soft gradient-sensitive loss can be defined as Equation \ref{eq:wglo}, in which $S_V$ is semantic labeling for $V$ and $S_R$ is semantic labeling for $R$.

\begin{equation} \label{eq:wglo}
\begin{aligned}
&L_{grad}(G_{V\rightarrow R},G_{R\rightarrow V},V,R,S_V,S_R,\alpha,\beta)\\
=&\mathbb{E}_{r\sim p_{data}(r)}[l_{s-grad}(r,s_r,G_{R\rightarrow V},\alpha,\beta)]\\
+& \mathbb{E}_{v\sim p_{data}(v)}[l_{s-grad}(v,s_v,G_{V\rightarrow R},\alpha,\beta)]
\end{aligned}
\end{equation}

\subsubsection{Full objective function}

\indent

Our full objective function is a combination of adversarial loss, cycle consistency loss and soft gradient-sensitive loss, as Equation \ref{eq:fo}, where $\lambda_c$ and $\lambda_g$ control the relative importance of cycle consistency loss and soft gradient-sensitive loss, compared with adversarial loss.

\begin{equation} \label{eq:fo}
\begin{aligned}
&L(G_{V\rightarrow R},G_{R\rightarrow V},\mathit{SD_V},\mathit{SD_R})\\
=&L_{adv}(G_{V\rightarrow R},\mathit{SD_R},V,R)\\
+&L_{adv}(G_{R\rightarrow V},\mathit{SD_V},R,V)\\
+&\lambda_c L_{cyc}(G_{V\rightarrow R},G_{R\rightarrow V},V,R)\\
+&\lambda_g L_{grad}(G_{V\rightarrow R},G_{R\rightarrow V},V,R,S_V,S_R,\alpha,\beta)\\
\end{aligned}
\end{equation}

Our optimization target can be then represented as:

\begin{equation} \label{eq:ot}
\begin{aligned}
&G_{V\rightarrow R}^*, G_{R\rightarrow V}^*  \\
=& \arg \min_{\substack{G_{V\rightarrow R}\\G_{R\rightarrow V}}} \max_{\substack{\mathit{SD_R}\\\mathit{SD_V}}} L(G_{V\rightarrow R},G_{R\rightarrow V},\mathit{SD_V},\mathit{SD_R})
\end{aligned}
\end{equation}

\subsection{Semantic-aware discriminator} \label{sec:sd}

\begin{figure}
\begin{tabular}{@{\hskip 0in}c@{\hskip 0.01in}c}
\includegraphics[width=0.499\linewidth]{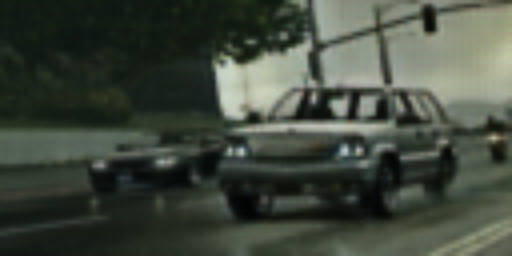} & \includegraphics[width=0.499\linewidth]{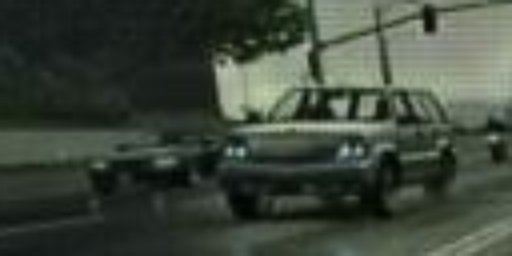}\\
\small{Adapted image with $L_{grad}$} & \small{Adapted image without $L_{grad}$} \\
\end{tabular}
   \caption{4X zoomed adapted images for showing the effectiveness of $L_{grad}$ objective.}\vspace{-6mm}
\label{fig:wgl-cmp}
\end{figure}

The introduction of soft gradient-sensitive loss contributes to smoother textures and clearer semantic boundaries (Figure \ref{fig:wgl-cmp}). However, the scene adaption also needs to retain more high-level semantic consistencies for each specific semantic region. A typical example is after the virtual-to-real adaption, the tone goes dark for the whole image as real-world images are not as luminous as virtual-world images, however, we may only want roads to be darker without changing much of the sky, or even make sky lighter. The reason for yielding such inappropriate holistic scene adaption is that the traditional discriminator only judges realism image-wise, regardless of texture differences in a semantic-aware manner. To make discriminator semantic-aware, we introduce semantic-aware discriminators $\mathit{SD_V}$ and $\mathit{SD_R}$. The idea is to create a separate channel for each different semantic class in the discriminator. In practice, this can be achieved by transiting the number of filters in the last layer of standard discriminator to number of semantic classes, and then applying semantic masks upon filters to let each of them focus on different semantic classes. 

More specifically, the last ($k$-th) layer's feature map of a standard discriminator is typically a tensor $\mathbf{T}_k$ with shape $(w_k,h_k,1)$, where $w_k$ stands for width and $h_k$ stands for height. $\mathbf{T}_k$ will then be compared with an all-one or all-zero tensor to calculate adversarial objective. In contrast, the semantic-aware discriminator we propose will change $\mathbf{T}_k$ as a tensor with shape $(w_k,h_k,s)$, where $s$ is the number of semantic classes. We then convert image's semantic labeling to one-hot style and resize to $(w_k,h_k)$, which will result in a mask $\mathbf{M}$ with same shape $(w_k,h_k,s)$, and $\{\mathbf{M}_{ij}\}\in \{0,1\}$. By multiplying $\mathbf{T}_k$ and $\mathbf{M}$ element-wise, each filter within $\mathbf{T}_k$ will only focus on one particular semantic class. Finally, by summing up $\mathbf{T}_k$ along the last dimension, a tensor with shape $(w_k,h_k,1)$ will be acquired and adversarial objective can be calculated the same way as the standard discriminator. Figure \ref{fig:dis} gives an illustration of proposed semantic-aware discriminator.

\begin{figure*}
\begin{center}
\includegraphics[width=1.0\linewidth]{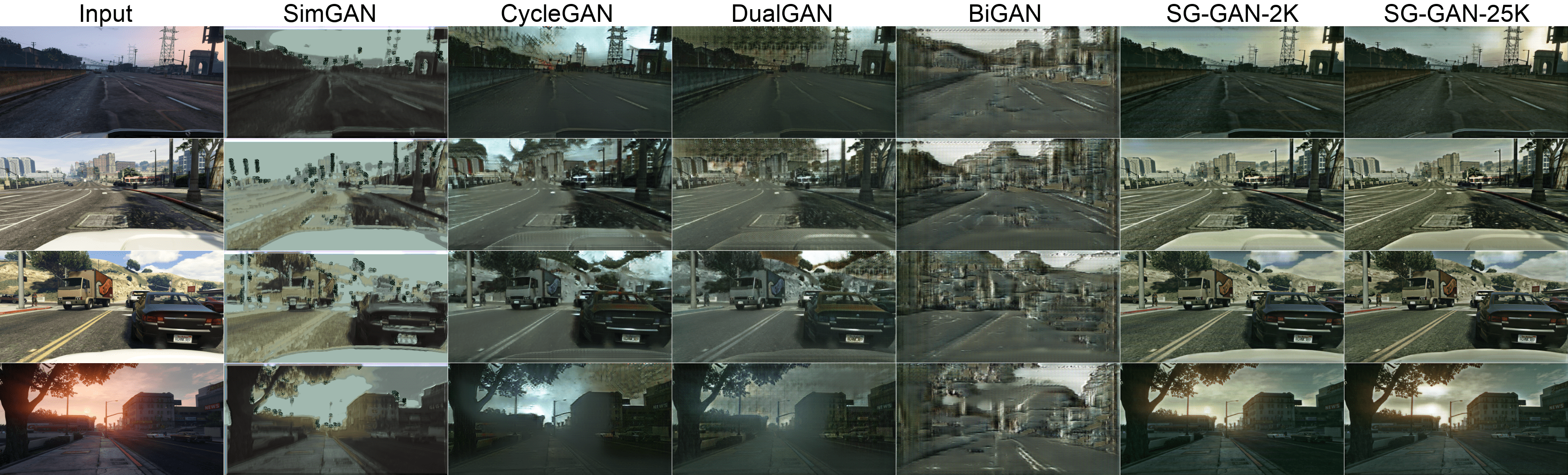}
\end{center}\vspace{-4mm}
   \caption{Visual comparison with state-of-the-art methods and our variants.}
\label{fig:vis}\vspace{-2mm}
\end{figure*}

\begin{table*}
\small
\begin{center}
\begin{tabular}{|l|c|c|c|c|c|c|}
\hline 
\diagbox{Method A}{Method B} & CycleGAN\cite{CycleGAN} & DualGAN\cite{dualgan} & SimGAN\cite{SimGAN} & BiGAN\cite{bigan} & SG-GAN-2K\\
\hline\hline
SG-GAN-2K & 79.2\% - 20.8\% & 93.4\% -  \phantom{0}6.6\% & 97.2\% - 2.8\% & 99.8\% - 0.2\% & ---\\
SG-GAN-25K & 83.4\% - 16.6\% & 94.0\% -  \phantom{0}6.0\% & 98.4\% - 1.6\% & 99.8\% - 0.2\% & 53.8\% - 46.2\% \\
\hline
\end{tabular}
\end{center}\vspace{-2mm}
\caption{Results of A/B tests on Amazon Mechanical Turk (AMT). Each cell compares the proportion that image adapted by one method is chosen as more realistic than the other, in the format of ``proportion of Method A - proportion of Method B".} \vspace{-6mm}
\label{table:amt}
\end{table*}

\begin{figure}
\begin{tabular}{@{\hskip 0in}c@{\hskip 0.01in}c}
\includegraphics[width=0.499\linewidth]{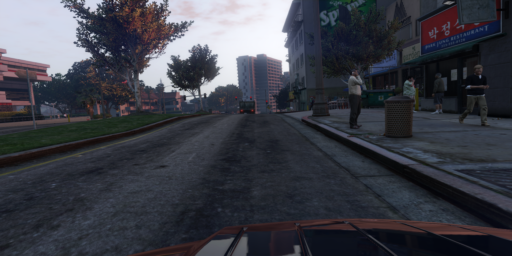} & \includegraphics[width=0.499\linewidth]{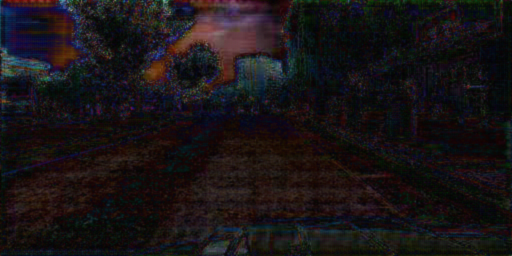}\\
\small{(a) Input} & \small{(b) Diff between (e) and (f)}\\
\includegraphics[width=0.499\linewidth]{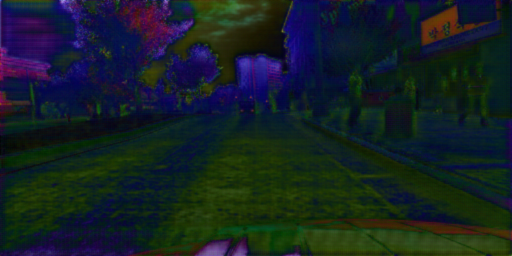} & \includegraphics[width=0.499\linewidth]{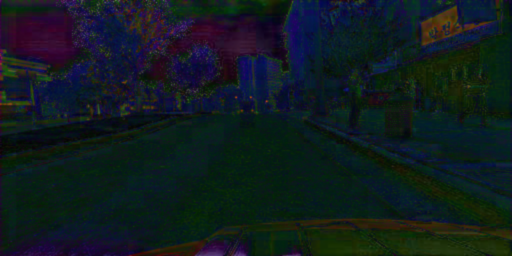}\\
\small{(c) Diff between (a) and (e)} & \small{(d) Diff between (a) and (f)}\\
\includegraphics[width=0.499\linewidth]{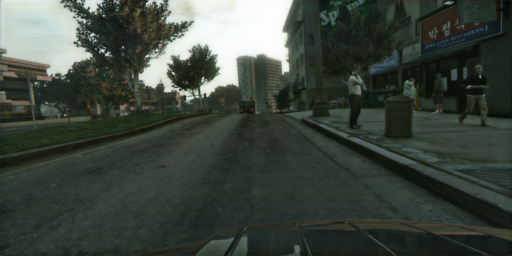} & \includegraphics[width=0.499\linewidth]{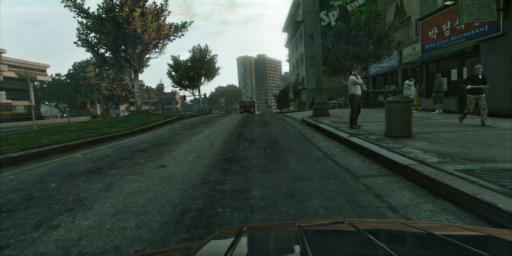}\\
\small{(e) Adapted image with $\mathit{SD}$} & \small{(f) Adapted image without $\mathit{SD}$}\\
\includegraphics[width=0.499\linewidth]{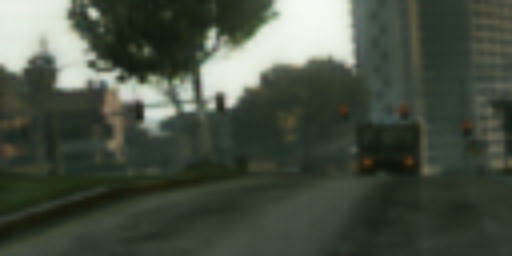} & \includegraphics[width=0.499\linewidth]{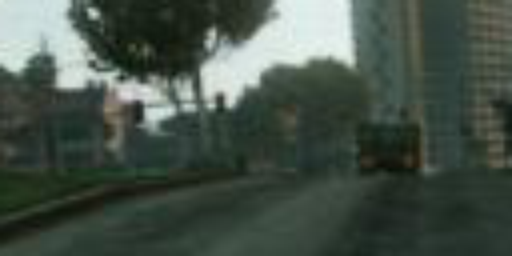}\\
\small{(g) 4X zoomed (e)} & \small{(h) 4X zoomed (f)} \\
\end{tabular}
   \caption{Comparison for showing the effectiveness of semantic-aware discriminator $\mathit{SD}$. (a) Input virtual-world image. (b) Absolute difference between (e) and (f). (c) Absolute difference between (a) and (e). (d) Absolute difference between (a) and (f). (e) Adapted image generated by SG-GAN-25K. (f) Adapted image generated by the variant without $\mathit{SD}$. (g) 4X zoomed details for (e). (h) 4X zoomed details for (f). Note how $\mathit{SD}$ helps with more diverse changes in tone and textures for different semantic classes by comparing (b),(c),(d). The comparison of (g) and (h) shows the ability of $\mathit{SD}$ to generate finer details, \eg, the faraway traffic light and smooth sky. }\vspace{-6mm}
\label{fig:semdis-cmp}
\end{figure}

\section{Experiments} \label{Experiments}

\subsection{Implementation} \label{Imple}

\textbf{Dataset.} We randomly sample 2000 images each from GTA-V dataset \cite{Playing} and Cityscapes training set \cite{Cityscapes} as training images for $V$ and $R$. Another 500 images each from GTA-V dataset and Cityscapes training set are sampled for visual comparison and validation. Cityscapes validation set is not used for validating adaption approaches here since it will later be applied to evaluate semantic segmentation scores in Section \ref{sec:exp-ss}. We train SG-GAN on such dataset and term it as \textbf{SG-GAN-2K}. The same dataset is used for training all baselines in Section \ref{sec:comparison}, making them comparable with SG-GAN-2K. To study the effect of virtual-world images, we further expand virtual-world training images to all 24966 images of GTA-V dataset, making a dataset with 24966 virtual images and 2000 real images. A variant of SG-GAN is trained on the expanded dataset and termed as \textbf{SG-GAN-25K}. 

\textbf{Network architecture.} We use $256\times 512$ images for training phase due to GPU memory limitation. For the generator, we adapt the architecture from Isola \etal \cite{pix2pix}, which is a U-Net structure with skip connections between low level and high level layers. For the semantic-aware discriminator, we use a variant of PatchGAN \cite{pix2pix,CycleGAN}, which is a fully convolutional network consists of multiple layers of (leaky-ReLU, instance norm \cite{instancenorm}, convolution) and helps the discriminator identify realism patch-wise. 

\textbf{Training details.} To stabilize training, we use history of refined images \cite{SimGAN} for training semantic-aware discriminators $\mathit{SD_V}$ and $\mathit{SD_R}$. Moreover, we apply least square objective instead of log likelihood objective for adversarial loss, which is shown helpful in stabilizing training and generating higher quality images, as proposed by Mao \etal \cite{lsgan}. For parameters in Equation \ref{eq:fo}, we set $\lambda_c=10$, $\lambda_g=5$. $(\alpha,\beta)$ is set as $(1,0)$ for the first three epochs and then changed to $(0.9,0.1)$. For gradient filters in Equation \ref{eq:wgl}, we use Sobel filter (Equation \ref{eq:sobel}) for $\mathbf{C}_i$ and filters in Equation \ref{eq:segfilter} for $\mathbf{C}_s$ to avoid artifacts on image borders caused by reflect padding. For number of semantic classes in semantic-aware discriminator, we cluster 30 classes \cite{Cityscapes} into 8 categories to avoid sparse classes, i.e., $s=8$. Learning rate is set as 0.0002 and we use a batch size of 1. We implement SG-GAN based on TensorFlow framework \cite{abadi2016tensorflow}, and train it with a single Nvidia GTX 1080.

\begin{equation} \label{eq:segfilter}
\begin{aligned}
C_x = \begin{pmatrix}
0 & 0 & 0\\
-1 & 0 & 1\\
0 & 0 & 0\\
\end{pmatrix}&,
C_y = \begin{pmatrix}
0 & 1 & 0\\
0 & 0 & 0\\
0 & -1 & 0\\
\end{pmatrix}\\
\end{aligned}
\end{equation}

\textbf{Testing.} Semantic information will only be needed at training time. At test time SG-GAN only requires images without semantic information. Since the generators and the discriminators we use are fully convolutional, SG-GAN can handle images with high resolution ($1024\times2048$) at test time. The testing time is 1.3 second/image with a single Nvidia GTX 1080.

\subsection{Comparison with state-of-the-art methods} \label{sec:comparison}
We compare our SG-GAN with current state-of-the-art baselines for unpaired virtual-to-real scene adaption for demonstrating its superiority. 

\subsubsection{Baselines}

\indent


\textbf{SimGAN} \cite{SimGAN} introduces a self-regularization for GAN and local adversarial loss to train a refiner for image adaption. In the experiments we use channel-wise mean values as self-regularization term. We use the architecture as proposed in \cite{SimGAN}. 

\textbf{CycleGAN} \cite{CycleGAN} learns mapping functions through adversarial loss and cycle consistency loss. It uses ResNet \cite{resnet} architecture for the generators and PatchGAN \cite{pix2pix} for the discriminators.

\textbf{DualGAN} \cite{dualgan} uses U-Net structure for generators that are identical with SG-GAN. It uses the same PatchGAN structure as CycleGAN, but different from CycleGAN it follows the loss format and training procedure proposed in Wasserstein GAN \cite{arjovsky2017wasserstein}.

\textbf{BiGAN} \cite{bigan,ALI} learns the inverse mapping of standard GANs \cite{GAN}. While standard GANs learn generators mapping random noises $Z$ to images $X$, i.e., $Z\to X$ , BiGAN \cite{bigan,ALI} also aims at inferring latent noises based on images $X\to Z$. By taking $Z$ as image, BiGAN can also be used for unpaired scene adaption. For the implementation of BiGAN we use the codes provided by \cite{CycleGAN}.

\subsubsection{Qualitative and quantitative evaluation}

\indent

Figure \ref{fig:vis} compares between SG-GAN-2K and other state-of-the-art methods visually. In general, SG-GAN generates better visualization results, in the form of clear boundaries, consistent semantic classes, smooth texture, etc. Moreover, SG-GAN-2K shows its ability for personalized adaption, \eg, while we retain the red color of vehicle's headlight, the red color of sunset is changed to sunny yellow that is closer to real-world images.

To further evaluate our approach quantitatively, we conduct A/B tests on Amazon Mechanical Turk (AMT) by comparing SG-GAN-2K and baseline approaches pairwise. We use 500 virtual-world images with size of $256\times 512$ as input, and present pairs of adapted images generated by different methods to workers for A/B tests. For each image-image pair, we ask workers which image is more realistic than the other and record their answers. There are 123 workers participated in our A/B tests and the results are shown in Table \ref{table:amt}. According to the statistics SG-GAN shows its superiority over all other approaches by a high margin. We attribute such superiority to clearer boundaries and smoother textures achieved by soft gradient-sensitive loss, and personalized texture rendering with the help of semantic-aware discriminator. 

\subsection{Ablation studies}\label{ablation}


\textbf{Effectiveness of soft gradient-sensitive objective.} To demonstrate the effectiveness of soft gradient-sensitive loss $L_{grad}$, we train a variant of SG-GAN without applying $L_{grad}$ and compare it with SG-GAN-25K. Figure \ref{fig:wgl-cmp} shows an example by inspecting details through a 4X zoom. Compared with SG-GAN-25K, the variant without $L_{grad}$ has coarse semantic boundaries and rough textures, which demonstrates soft gradient-sensitive loss can help generate adapted images with clearer semantic boundaries and smoother textures. 

\textbf{Effectiveness of semantic-aware discriminator.} We use a variant of SG-GAN without applying semantic-aware discriminator ($\mathit{SD}$) and compare it with SG-GAN-25K to study the effectiveness of $\mathit{SD}$. As shown in Figure \ref{fig:semdis-cmp}, comparing (g) and (h), the variant without $\mathit{SD}$ lacks for details, \eg, the color of traffic light, and generates coarser textures, \eg, the sky. The difference maps, i.e., (b), (c), (d) in Figure \ref{fig:semdis-cmp}, further reveal that semantic-aware discriminator leads to personalized texture rendering for each distinct region with specific semantic meaning.


\textbf{The effect of virtual training image size.} Figure \ref{fig:vis} compares variants of SG-GAN that use distinct numbers of virtual-world images for training. Generally, SG-GAN-25K generates clearer details than SG-GAN-2K for some images. Further A/B tests between them in Table \ref{table:amt} show SG-GAN-25K is slightly better than SG-GAN-2K because of using more training data. Both qualitative and quantitative comparisons indicate more data could help, however, the improved performance may be only notable if dataset difference is in orders of magnitude.

\begin{figure}
\begin{tabular}{@{\hskip 0in}c@{\hskip 0.01in}c}
\includegraphics[width=0.499\linewidth]{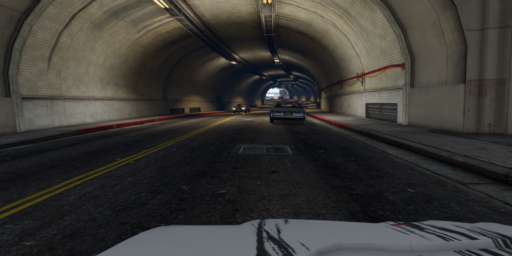} & \includegraphics[width=0.499\linewidth]{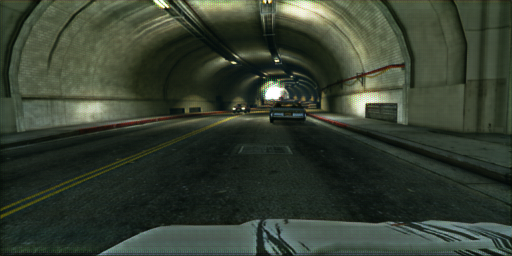}\\
\small{(a) Input} & \small{(b) Adapted}\\
\end{tabular}
   \caption{A very rare unsatisfactory example. (a) Input virtual-world image. (b) Adapted image generated by SG-GAN.}\vspace{-2mm}
\label{fig:unsat}
\end{figure}

\textbf{Discussion.} \label{sec:discuss}
While SG-GAN generates realistic results for almost all tested images, in very rare case the adapted image is unsatisfactory, as shown in Figure \ref{fig:unsat}. Our model learns the existence of sunlight, however it is unaware of the image is taken in a tunnel and thus sunlight would be abnormal. We attribute this rare unsatisfactory case to the lack of diversity of real-world dataset compared with virtual-world dataset, thus such case could be seen as an outlier.

More real-world images could help alleviate such unsatisfactory case, but SG-GAN is restricted by the limited number of fine-grained real-world annotations for training, \eg, Cityscapes dataset only contains a fine-grained training set of 2975 images. However, we foresee a possibility to solve the data insufficient issue by using coarse annotations labeled by human or semantic segmentation models. In our implementation of semantic-aware discriminator, semantic masks are actually clustered to avoid sparse classes, \eg, semantic classes ``building", ``wall", ``fence", ``guard rail", ``bridge" and ``tunnel" are clustered into a single mask indicating ``construction". Considering such cluster, annotation granularity may not be a vital factor for our model. Thus investigating the trade-off between annotation granularity and dataset size would be a possible next step.

\subsection{Application on semantic segmentation} \label{sec:exp-ss}
To further demonstrate the scene adaption quality of SG-GAN, we conduce comparisons on the downstream semantic segmentation task on Cityscapes validation set \cite{Cityscapes} by adapting from GTA-V dataset \cite{Playing}, similar to \cite{hoffman2016fcns}. The idea is to train semantic segmentation model merely based on adapted virtual-world data, i.e., 24966 images of GTA-V dataset \cite{Playing}, and evaluate model's performance on real-world data, i.e., Cityscapes validation set \cite{Cityscapes}. For the semantic segmentation model we use the architecture proposed by Wu \etal \cite{ademxapp} and exactly follow its training procedure, which shows impressive results on Cityscapes dataset. Table \ref{table:ss} shows the results. The baseline method is the version that trains semantic segmentation model directly on original virtual-world data and groundtruth pairs. 

\begin{figure}
\begin{tabular}{@{\hskip 0in}c@{\hskip 0.01in}c}
\includegraphics[width=0.499\linewidth]{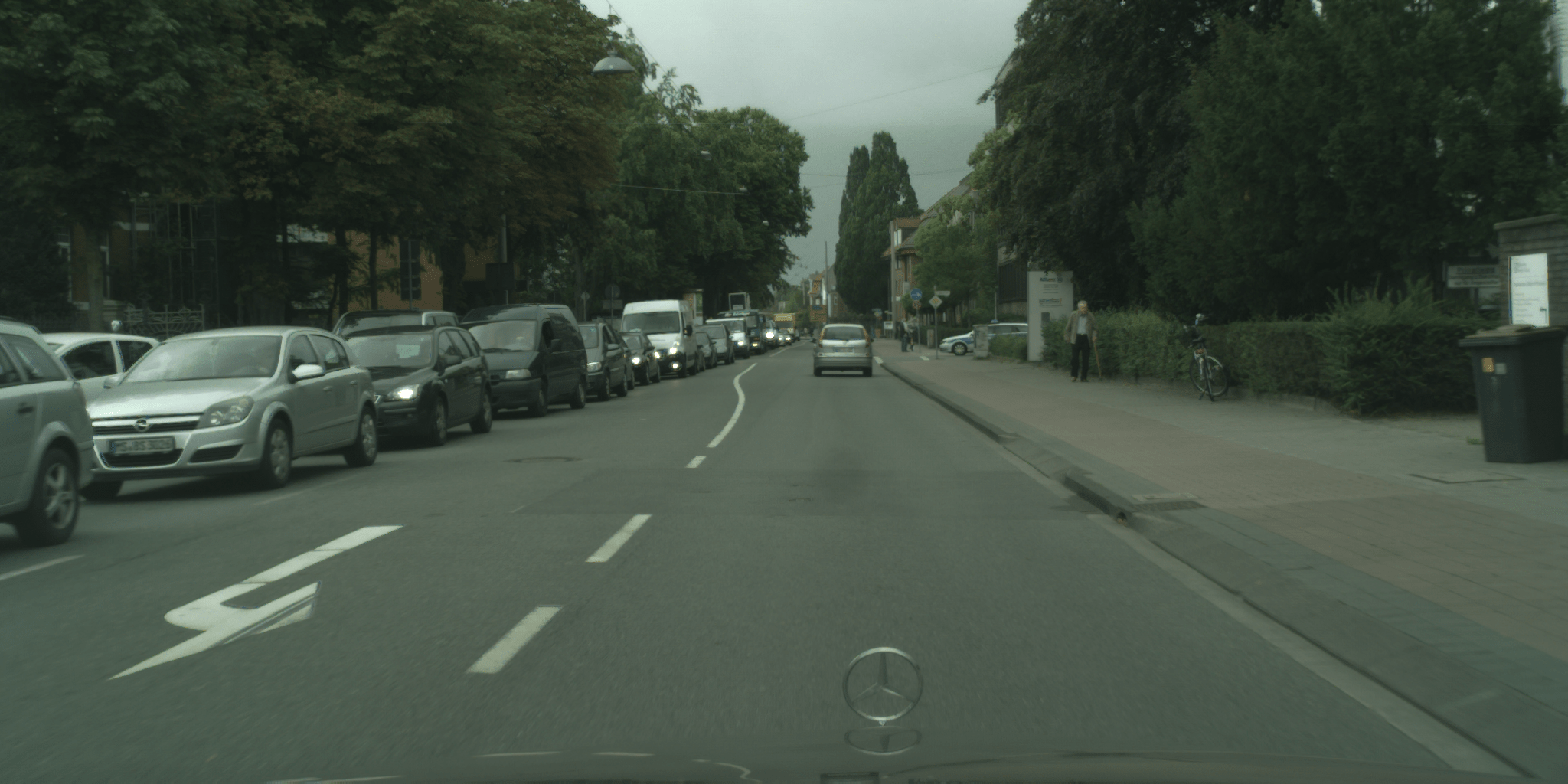} & \includegraphics[width=0.499\linewidth]{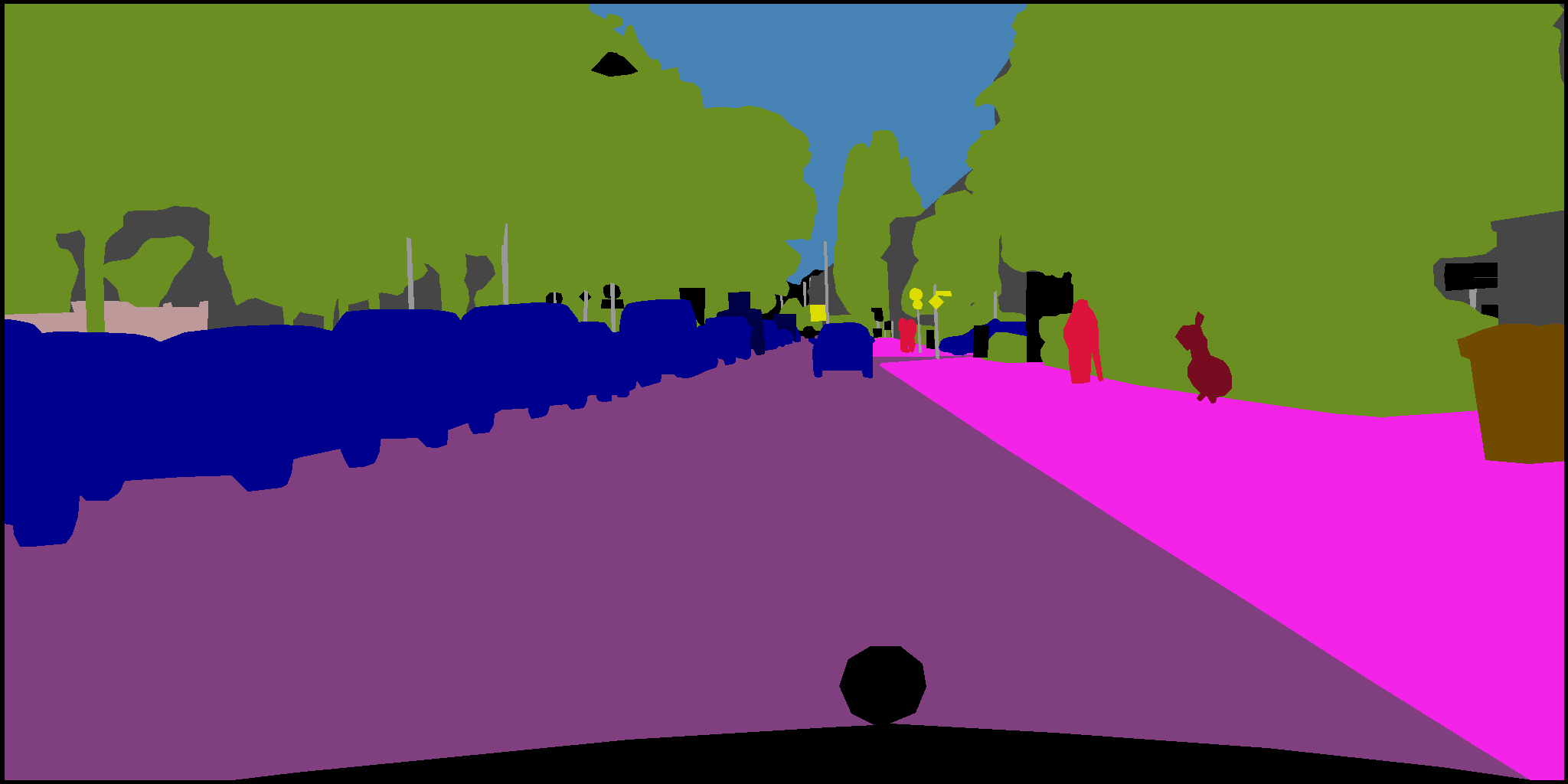}\\
\small{(a) Real-world image} & \small{(b) Groundtruth} \\
\includegraphics[width=0.499\linewidth]{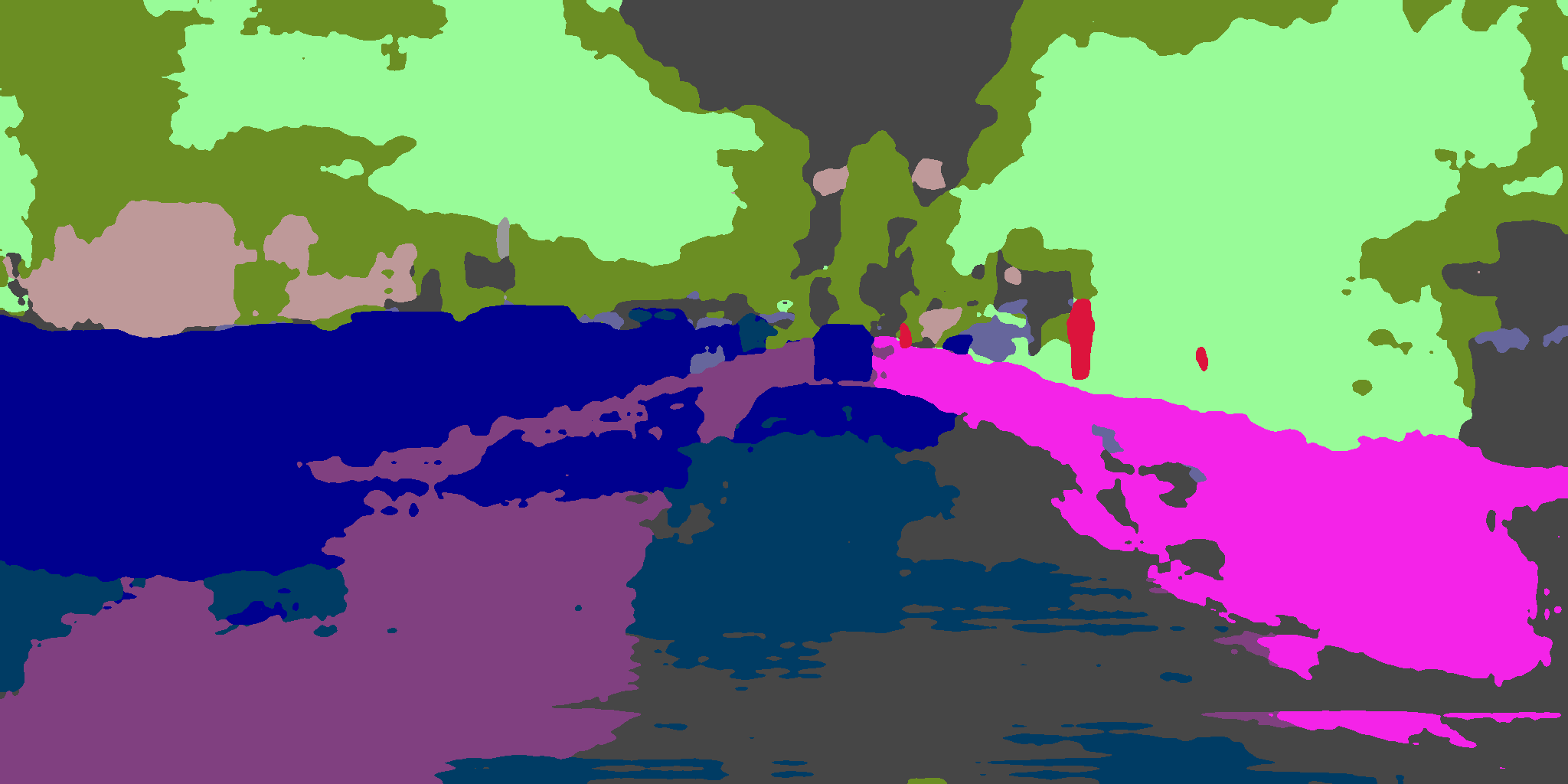} & \includegraphics[width=0.499\linewidth]{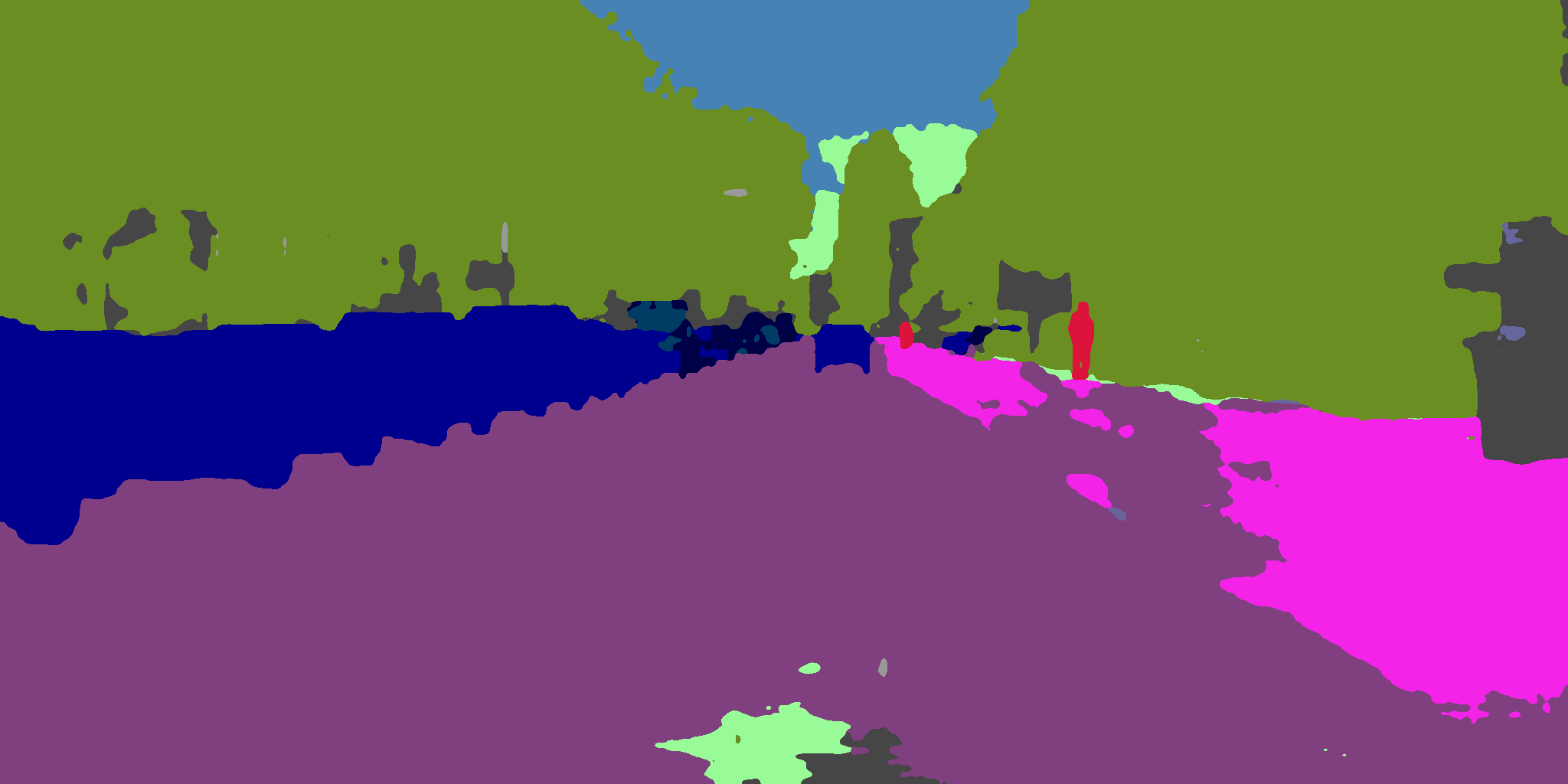}\\
\small{(c) Baseline} & \small{(d) SG-GAN-25K} \\
\end{tabular}
   \caption{Comparison of segmentation results. Color scheme of segmentation is the same as Cityscapes \cite{Cityscapes}.}\vspace{-6mm}
\label{fig:seg-cmp}
\end{figure}

We first compare SG-GAN with CycleGAN \cite{CycleGAN}. The substantially higher semantic segmentation performance by SG-GAN shows its ability to yield adapted images closer to real-world data distribution. Figure \ref{fig:seg-cmp} illustrates the visual comparison between SG-GAN and baseline to further show how SG-GAN helps improve segmentation. We further compare our approach with a hidden feature representation based adaption method proposed by Huffman \etal \cite{hoffman2016fcns}, and SG-GAN achieves a high performance margin. These evaluations on semantic segmentation again confirm SG-GAN's ability to adapt high quality images, benefiting from preserving consistent semantic information and rendering personalized texture closer to real-world via soft gradient-sensitive objective and semantic discriminator.

\begin{table}
\small
\begin{center}
\begin{tabular}{|l|c|c|c|}
\hline
Method & Pixel acc. & Class acc. & Class IOU \\
\hline\hline
Baseline & 54.51 & 35.95 & 24.60\\
Hoffman \etal \cite{hoffman2016fcns} & -- & -- & 27.10 \\
CycleGAN\cite{CycleGAN} & 71.61 & 42.98 & 28.15\\
SG-GAN-2K & 72.65 & 45.87 & 33.81\\
SG-GAN-25K & \textbf{81.72} & \textbf{47.29} & \textbf{37.43}\\
\hline
\end{tabular}
\end{center}\vspace{-2mm}
\caption{Comparison of semantic segmentation scores (\%) on Cityscapes 500 images validation set.}\vspace{-2mm}
\label{table:ss}
\end{table}

\section{Conclusion} \label{Conclusion}
In this work, we propose a novel SG-GAN for virtual-to-real urban scene adaption with the good property of retaining critical semantic information. SG-GAN employs a new soft gradient-sensitive loss to confine clear semantic boundaries and smooth adapted texture, and a semantic-aware discriminator to personalize texture rendering. We conduct extensive experiments to compare SG-GAN with other state-of-the-art domain adaption approaches both qualitatively and quantitatively, which all demonstrate the superiority of SG-GAN. Further experiments on the downstream semantic segmentation confirm the effectiveness of SG-GAN in virtual-to-real urban scene adaption. In future, we plan to apply our model on Playing-for-Benchmarks \cite{PlayingForBenchmarks} dataset, which has an order of magnitude more annotated data from virtual-world for further boosting adaption performance.

{\small
\bibliographystyle{ieee}
\bibliography{egbib}

\begin{thebibliography}{10}\itemsep=-1pt

\bibitem{abadi2016tensorflow}
M.~Abadi, A.~Agarwal, P.~Barham, E.~Brevdo, Z.~Chen, C.~Citro, G.~S. Corrado,
  A.~Davis, J.~Dean, M.~Devin, et~al.
\newblock Tensorflow: Large-scale machine learning on heterogeneous distributed
  systems.
\newblock {\em arXiv preprint arXiv:1603.04467}, 2016.

\bibitem{arjovsky2017wasserstein}
M.~Arjovsky, S.~Chintala, and L.~Bottou.
\newblock Wasserstein gan.
\newblock {\em arXiv preprint arXiv:1701.07875}, 2017.

\bibitem{bansal2017pixelnn}
A.~Bansal, Y.~Sheikh, and D.~Ramanan.
\newblock Pixelnn: Example-based image synthesis.
\newblock {\em arXiv preprint arXiv:1708.05349}, 2017.

\bibitem{CamVid}
G.~J. Brostow, J.~Fauqueur, and R.~Cipolla.
\newblock Semantic object classes in video: A high-definition ground truth
  database.
\newblock {\em Pattern Recognition Letters}, 2009.

\bibitem{cha2017adversarial}
M.~Cha, Y.~Gwon, and H.~Kung.
\newblock Adversarial nets with perceptual losses for text-to-image synthesis.
\newblock {\em arXiv preprint arXiv:1708.09321}, 2017.

\bibitem{CRN}
Q.~Chen and V.~Koltun.
\newblock Photographic image synthesis with cascaded refinement networks.
\newblock In {\em ICCV}, 2017.

\bibitem{Cityscapes}
M.~Cordts, M.~Omran, S.~Ramos, T.~Rehfeld, M.~Enzweiler, R.~Benenson,
  U.~Franke, S.~Roth, and B.~Schiele.
\newblock The cityscapes dataset for semantic urban scene understanding.
\newblock In {\em CVPR}, 2016.

\bibitem{bigan}
J.~Donahue, P.~Kr{\"a}henb{\"u}hl, and T.~Darrell.
\newblock Adversarial feature learning.
\newblock {\em arXiv preprint arXiv:1605.09782}, 2016.

\bibitem{ALI}
V.~Dumoulin, I.~Belghazi, B.~Poole, A.~Lamb, M.~Arjovsky, O.~Mastropietro, and
  A.~Courville.
\newblock Adversarially learned inference.
\newblock {\em arXiv preprint arXiv:1606.00704}, 2016.

\bibitem{NeuralStyle}
L.~Gatys, A.~Ecker, and M.~Bethge.
\newblock A neural algorithm of artistic style.
\newblock {\em Nature Communications}, 2015.

\bibitem{gatys2016preserving}
L.~A. Gatys, M.~Bethge, A.~Hertzmann, and E.~Shechtman.
\newblock Preserving color in neural artistic style transfer.
\newblock {\em arXiv preprint arXiv:1606.05897}, 2016.

\bibitem{Gebru_2017_ICCV}
T.~Gebru, J.~Hoffman, and L.~Fei-Fei.
\newblock Fine-grained recognition in the wild: A multi-task domain adaptation
  approach.
\newblock In {\em ICCV}, 2017.

\bibitem{GAN}
I.~Goodfellow, J.~Pouget-Abadie, M.~Mirza, B.~Xu, D.~Warde-Farley, S.~Ozair,
  A.~Courville, and Y.~Bengio.
\newblock Generative adversarial nets.
\newblock In {\em NIPS}, 2014.

\bibitem{resnet}
K.~He, X.~Zhang, S.~Ren, and J.~Sun.
\newblock Deep residual learning for image recognition.
\newblock In {\em CVPR}, 2016.

\bibitem{hoffman2014lsda}
J.~Hoffman, S.~Guadarrama, E.~S. Tzeng, R.~Hu, J.~Donahue, R.~Girshick,
  T.~Darrell, and K.~Saenko.
\newblock Lsda: Large scale detection through adaptation.
\newblock In {\em NIPS}, 2014.

\bibitem{hoffman2015detector}
J.~Hoffman, D.~Pathak, T.~Darrell, and K.~Saenko.
\newblock Detector discovery in the wild: Joint multiple instance and
  representation learning.
\newblock In {\em CVPR}, 2015.

\bibitem{hoffman2016large}
J.~Hoffman, D.~Pathak, E.~Tzeng, J.~Long, S.~Guadarrama, T.~Darrell, and
  K.~Saenko.
\newblock Large scale visual recognition through adaptation using joint
  representation and multiple instance learning.
\newblock {\em JMLR}, 2016.

\bibitem{hoffman2016fcns}
J.~Hoffman, D.~Wang, F.~Yu, and T.~Darrell.
\newblock Fcns in the wild: Pixel-level adversarial and constraint-based
  adaptation.
\newblock {\em arXiv preprint arXiv:1612.02649}, 2016.

\bibitem{hu2017squeeze}
J.~Hu, L.~Shen, and G.~Sun.
\newblock Squeeze-and-excitation networks.
\newblock {\em arXiv preprint arXiv:1709.01507}, 2017.

\bibitem{huang2017densely}
G.~Huang, Z.~Liu, L.~van~der Maaten, and K.~Q. Weinberger.
\newblock Densely connected convolutional networks.
\newblock In {\em CVPR}, 2017.

\bibitem{pix2pix}
P.~Isola, J.-Y. Zhu, T.~Zhou, and A.~A. Efros.
\newblock Image-to-image translation with conditional adversarial networks.
\newblock {\em arXiv preprint arXiv:1611.07004}, 2016.

\bibitem{perceptual}
J.~Johnson, A.~Alahi, and L.~Fei-Fei.
\newblock Perceptual losses for real-time style transfer and super-resolution.
\newblock In {\em ECCV}, 2016.

\bibitem{Driving}
M.~Johnson-Roberson, C.~Barto, R.~Mehta, S.~N. Sridhar, K.~Rosaen, and
  R.~Vasudevan.
\newblock Driving in the matrix: Can virtual worlds replace human-generated
  annotations for real world tasks?
\newblock In {\em ICRA}, 2017.

\bibitem{discogan}
T.~Kim, M.~Cha, H.~Kim, J.~Lee, and J.~Kim.
\newblock Learning to discover cross-domain relations with generative
  adversarial networks.
\newblock {\em arXiv preprint arXiv:1703.05192}, 2017.

\bibitem{MMD}
Y.~Li, N.~Wang, J.~Liu, and X.~Hou.
\newblock Demystifying neural style transfer.
\newblock {\em arXiv preprint arXiv:1701.01036}, 2017.

\bibitem{liang2017recurrent}
X.~Liang, Z.~Hu, H.~Zhang, C.~Gan, and E.~P. Xing.
\newblock Recurrent topic-transition gan for visual paragraph generation.
\newblock {\em arXiv preprint arXiv:1703.07022}, 2017.

\bibitem{liang2017dual}
X.~Liang, L.~Lee, W.~Dai, and E.~P. Xing.
\newblock Dual motion gan for future-flow embedded video prediction.
\newblock In {\em IEEE International Conference on Computer Vision (ICCV)},
  volume~1, 2017.

\bibitem{liang2017generative}
X.~Liang, H.~Zhang, and E.~P. Xing.
\newblock Generative semantic manipulation with contrasting gan.
\newblock {\em arXiv preprint arXiv:1708.00315}, 2017.

\bibitem{unit}
M.-Y. Liu, T.~Breuel, and J.~Kautz.
\newblock Unsupervised image-to-image translation networks.
\newblock {\em arXiv preprint arXiv:1703.00848}, 2017.

\bibitem{liu2016coupled}
M.-Y. Liu and O.~Tuzel.
\newblock Coupled generative adversarial networks.
\newblock In {\em NIPS}, 2016.

\bibitem{lsgan}
X.~Mao, Q.~Li, H.~Xie, R.~Y. Lau, and Z.~Wang.
\newblock Multi-class generative adversarial networks with the l2 loss
  function.
\newblock {\em arXiv preprint arXiv:1611.04076}, 2016.

\bibitem{quionero2009dataset}
J.~Quionero-Candela, M.~Sugiyama, A.~Schwaighofer, and N.~D. Lawrence.
\newblock {\em Dataset shift in machine learning}.
\newblock The MIT Press, 2009.

\bibitem{reed2016generative}
S.~Reed, Z.~Akata, X.~Yan, L.~Logeswaran, B.~Schiele, and H.~Lee.
\newblock Generative adversarial text to image synthesis.
\newblock In {\em ICML}, 2016.

\bibitem{ren2015faster}
S.~Ren, K.~He, R.~Girshick, and J.~Sun.
\newblock Faster r-cnn: Towards real-time object detection with region proposal
  networks.
\newblock In {\em NIPS}, 2015.

\bibitem{PlayingForBenchmarks}
S.~R. Richter, Z.~Hayder, and V.~Koltun.
\newblock Playing for benchmarks.
\newblock In {\em ICCV}, 2017.

\bibitem{Playing}
S.~R. Richter, V.~Vineet, S.~Roth, and V.~Koltun.
\newblock Playing for data: {G}round truth from computer games.
\newblock In {\em ECCV}, 2016.

\bibitem{SimGAN}
A.~Shrivastava, T.~Pfister, O.~Tuzel, J.~Susskind, W.~Wang, and R.~Webb.
\newblock Learning from simulated and unsupervised images through adversarial
  training.
\newblock In {\em CVPR}, 2017.

\bibitem{Sobel}
I.~Sobel.
\newblock An isotropic 3$\times$ 3 image gradient operator.
\newblock {\em Machine vision for three-dimensional scenes}, 1990.

\bibitem{Adda_CVPR2017}
E.~Tzeng, J.~Hoffman, T.~Darrell, and K.~Saenko.
\newblock Adversarial discriminative domain adaptation.
\newblock In {\em CVPR}, 2017.

\bibitem{ulyanov2016texture}
D.~Ulyanov, V.~Lebedev, A.~Vedaldi, and V.~Lempitsky.
\newblock Texture networks: feed-forward synthesis of textures and stylized
  images.
\newblock In {\em ICML}, 2016.

\bibitem{instancenorm}
D.~Ulyanov, A.~Vedaldi, and V.~Lempitsky.
\newblock Instance normalization: The missing ingredient for fast stylization.
\newblock {\em arXiv preprint arXiv:1607.08022}, 2016.

\bibitem{ademxapp}
Z.~Wu, C.~Shen, and A.~v.~d. Hengel.
\newblock Wider or deeper: Revisiting the resnet model for visual recognition.
\newblock {\em arXiv preprint arXiv:1611.10080}, 2016.

\bibitem{Curse}
J.~Xie, M.~Kiefel, M.-T. Sun, and A.~Geiger.
\newblock Semantic instance annotation of street scenes by 3d to 2d label
  transfer.
\newblock In {\em CVPR}, 2016.

\bibitem{xiong2017learning}
W.~Xiong, W.~Luo, L.~Ma, W.~Liu, and J.~Luo.
\newblock Learning to generate time-lapse videos using multi-stage dynamic
  generative adversarial networks.
\newblock {\em arXiv preprint arXiv:1709.07592}, 2017.

\bibitem{dualgan}
Z.~Yi, H.~Zhang, P.~Tan, and M.~Gong.
\newblock Dualgan: Unsupervised dual learning for image-to-image translation.
\newblock In {\em ICCV}, 2017.

\bibitem{han2017stackgan}
H.~Zhang, T.~Xu, H.~Li, S.~Zhang, X.~Wang, X.~Huang, and D.~Metaxas.
\newblock Stackgan: Text to photo-realistic image synthesis with stacked
  generative adversarial networks.
\newblock In {\em ICCV}, 2017.

\bibitem{CycleGAN}
J.-Y. Zhu, T.~Park, P.~Isola, and A.~A. Efros.
\newblock Unpaired image-to-image translation using cycle-consistent
  adversarial networks.
\newblock In {\em ICCV}, 2017.

\end{thebibliography}
}

\end{document}